\documentclass[letterpaper]{article}
\usepackage{aaai}
\usepackage{times}
\usepackage{helvet}
\usepackage{courier}
\usepackage{color,xcolor}
\usepackage{epsfig}
\usepackage{graphicx}

\usepackage{tikz}
\usetikzlibrary{arrows}

\usepackage{tabularx}
\usepackage{adjustbox}
\usepackage{array}
\usepackage{booktabs}
\usepackage{colortbl}
\usepackage{float,wrapfig}
\usepackage{hhline}
\usepackage{multirow}
\usepackage{subcaption} %

\usepackage{amsmath,amsfonts,amsthm,amssymb}
\usepackage{bm}
\usepackage{nicefrac}
\usepackage{microtype}
\usepackage{dsfont}
\usepackage{changepage}
\usepackage{extramarks}
\usepackage{fancyhdr}
\usepackage{lastpage}
\usepackage{setspace}
\usepackage{soul}
\usepackage{xspace}

\usepackage[pagebackref=true,breaklinks=true,colorlinks,citecolor=gray]{hyperref}
\usepackage{url}

\usepackage{algorithm, algorithmic}
\usepackage{enumitem}
\usepackage{todonotes} %
\usepackage{natbib} %
\usepackage{titlesec}

\usepackage{amsmath,amsfonts,bm}

\def\rvc{{\mathbf{c}}}

\def\rvr{{\mathbf{r}}}
\def\rvs{{\mathbf{s}}}

\def\rvx{{\mathbf{x}}}

\def\rvz{{\mathbf{z}}}

\def\vzero{{\bm{0}}}

\DeclareMathAlphabet{\mathsfit}{\encodingdefault}{\sfdefault}{m}{sl}
\SetMathAlphabet{\mathsfit}{bold}{\encodingdefault}{\sfdefault}{bx}{n}

\def\sP{{\mathbb{P}}}

\def\sU{{\mathbb{U}}}

\usepackage{amsmath,amsfonts,bm}

\newcommand{\fid}{Fr\'echet Inception Distance\xspace}

\newcommand{\pgan}{Progressive GANs\xspace}

\newcommand{\pixel}{\text{P}}
\newcommand{\pixelall}{\sP}

\newcommand{\U}{\text{U}}
\newcommand{\Uall}{\sU}

\newcommand{\thresU}{\rvc}

\newcommand{\rUP}{\repr_{\U, \pixel}}
\newcommand{\rUPb}{\repr_{\overline{\U, \pixel}}}

\newcommand{\Ezp}{\E_{\rvz,\pixel}}
\newcommand{\IoU}{\text{IoU}}
\newcommand{\ACE}{\delta}
\newcommand{\ACEUc}{\ACE_{\U\rightarrow c}}
\newcommand{\xinsert}{\rvx_{i}}
\newcommand{\xablate}{\rvx_{a}}
\newcommand{\layer}[1]{\texttt{layer#1\xspace}}

\newcommand{\G}{G}
\newcommand{\seg}{\rvs_c}
\newcommand{\repr}{\rvr}

\newcommand{\uprepru}{\repr_{u, \sP}^{\uparrow}}
\newcommand{\f}{f}

\newcommand{\reffig}[1]{Figure~\ref{fig:#1}}

\newcommand{\reftbl}[1]{Table~\ref{tbl:#1}}

\newcommand{\lblfig}[1]{\label{fig:#1}}
\newcommand{\lblsec}[1]{\label{sec:#1}}

\newcommand{\lbltbl}[1]{\label{tbl:#1}}

\newcommand{\ignorethis}[1]{}

\newcommand{\myparagraph}[1]{\paragraph{#1}}

\def\eqref#1{equation~\ref{#1}}

\def\1{\bm{1}}

\def\rvc{{\mathbf{c}}}

\def\rvr{{\mathbf{r}}}
\def\rvs{{\mathbf{s}}}

\def\rvx{{\mathbf{x}}}

\def\rvz{{\mathbf{z}}}

\def\vzero{{\bm{0}}}

\DeclareMathAlphabet{\mathsfit}{\encodingdefault}{\sfdefault}{m}{sl}
\SetMathAlphabet{\mathsfit}{bold}{\encodingdefault}{\sfdefault}{bx}{n}

\def\sP{{\mathbb{P}}}

\def\sU{{\mathbb{U}}}

\newcommand{\E}{\mathbb{E}}

\DeclareMathOperator*{\argmax}{arg\,max}

\newcolumntype{L}[1]{>{\raggedright\let\newline\\\arraybackslash\hspace{0pt}}m{#1}}
\newcolumntype{C}[1]{>{\centering\let\newline\\\arraybackslash\hspace{0pt}}m{#1}}
\newcolumntype{R}[1]{>{\raggedleft\let\newline\\\arraybackslash\hspace{0pt}}m{#1}}

\newcommand{\ignore}[1]{}

\makeatletter
\DeclareRobustCommand\onedot{\futurelet\@let@token\@onedot}
\def\@onedot{\ifx\@let@token.\else.\null\fi\xspace}

\makeatother

\definecolor{MyDarkBlue}{rgb}{0,0.08,1}
\definecolor{MyDarkGreen}{rgb}{0.02,0.6,0.02}
\definecolor{MyDarkRed}{rgb}{0.8,0.02,0.02}
\definecolor{MyDarkOrange}{rgb}{0.40,0.2,0.02}
\definecolor{MyPurple}{RGB}{111,0,255}
\definecolor{MyRed}{rgb}{1.0,0.0,0.0}
\definecolor{MyGold}{rgb}{0.75,0.6,0.12}
\definecolor{MyDarkgray}{rgb}{0.66, 0.66, 0.66}

\begin{document}
\title{On the Units of GANs}

\author{David Bau\textsuperscript{1},
Jun-Yan Zhu\textsuperscript{1},
Hendrik Strobelt\textsuperscript{2},
Bolei Zhou\textsuperscript{3}, \\
{\bf \Large Joshua B. Tenenbaum\textsuperscript{1},
William T. Freeman\textsuperscript{1},
Antonio Torralba\textsuperscript{1}} \\
\textsuperscript{1}Massachusetts Institute of Technology,
\textsuperscript{2}IBM Research, Cambridge MA,
\textsuperscript{3} The Chinese University of Hong Kong \\
\textsuperscript{1}[davidbau, junyanz, jbt, billf, torralba]@csail.mit.edu, \textsuperscript{2}hendrik.strobelt@ibm.com 
\textsuperscript{3}bzhou@ie.cuhk.edu.hk }

\nocopyright
\maketitle

\global\csname @topnum\endcsname 0
\global\csname @botnum\endcsname 0
\renewcommand{\dbltopfraction}{1.0}
\renewcommand{\textfraction}{0}
\renewcommand{\dblfloatpagefraction}{0.5}

\begin{figure}[t]
\centering
\includegraphics[width=1.0\columnwidth]{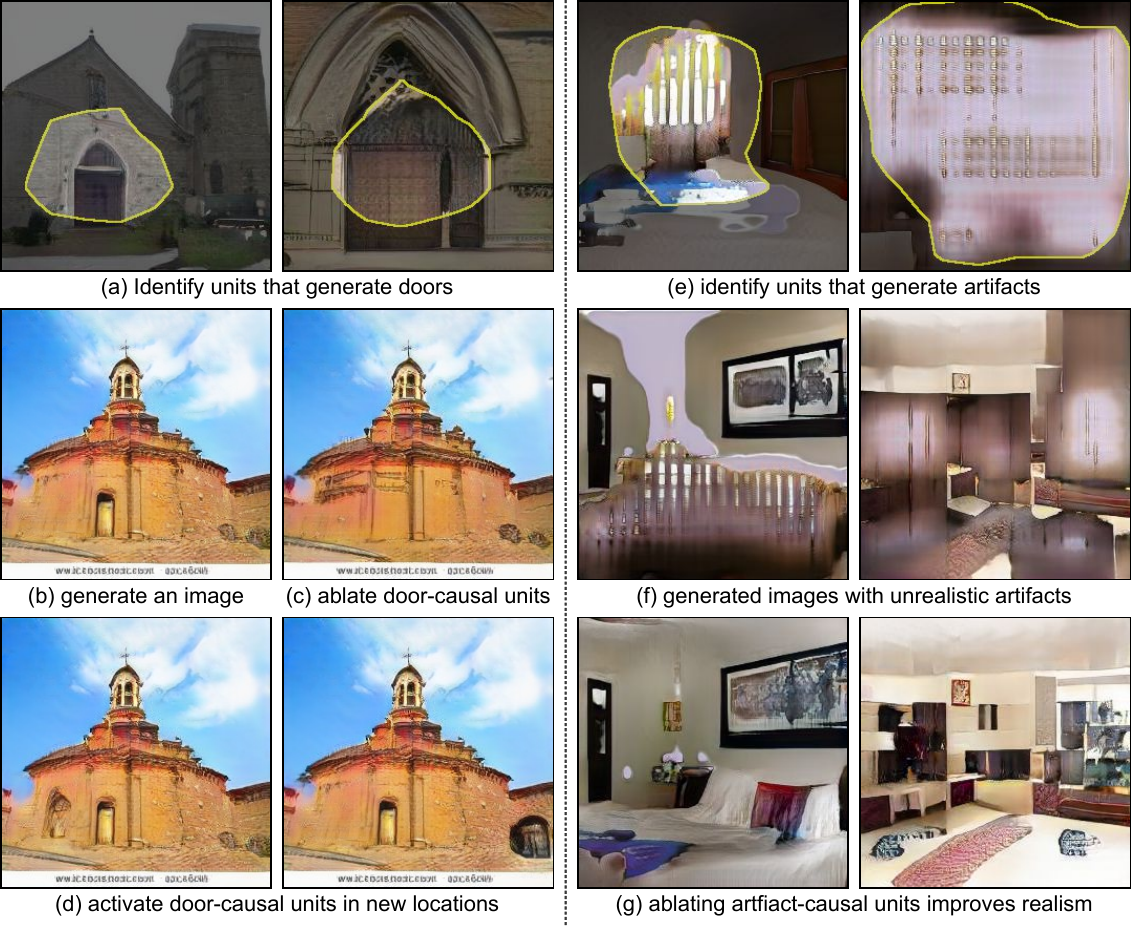}
\vspace{-20pt}
\caption{Overview: (a-d) We analyze how internal representations relate to (b) output of a Progressive GAN by identifying (a) units that correlate with object concepts (here doors) and (c) intervening in those units to remove and (d) add objects.  (e-g) Our framework can be used to (e) identify units that (f) cause artifacts and (g) reduce artifacts when ablated.}  
\lblfig{teaser}
\end{figure}

\begin{figure}
\centering
\includegraphics[width=\columnwidth]{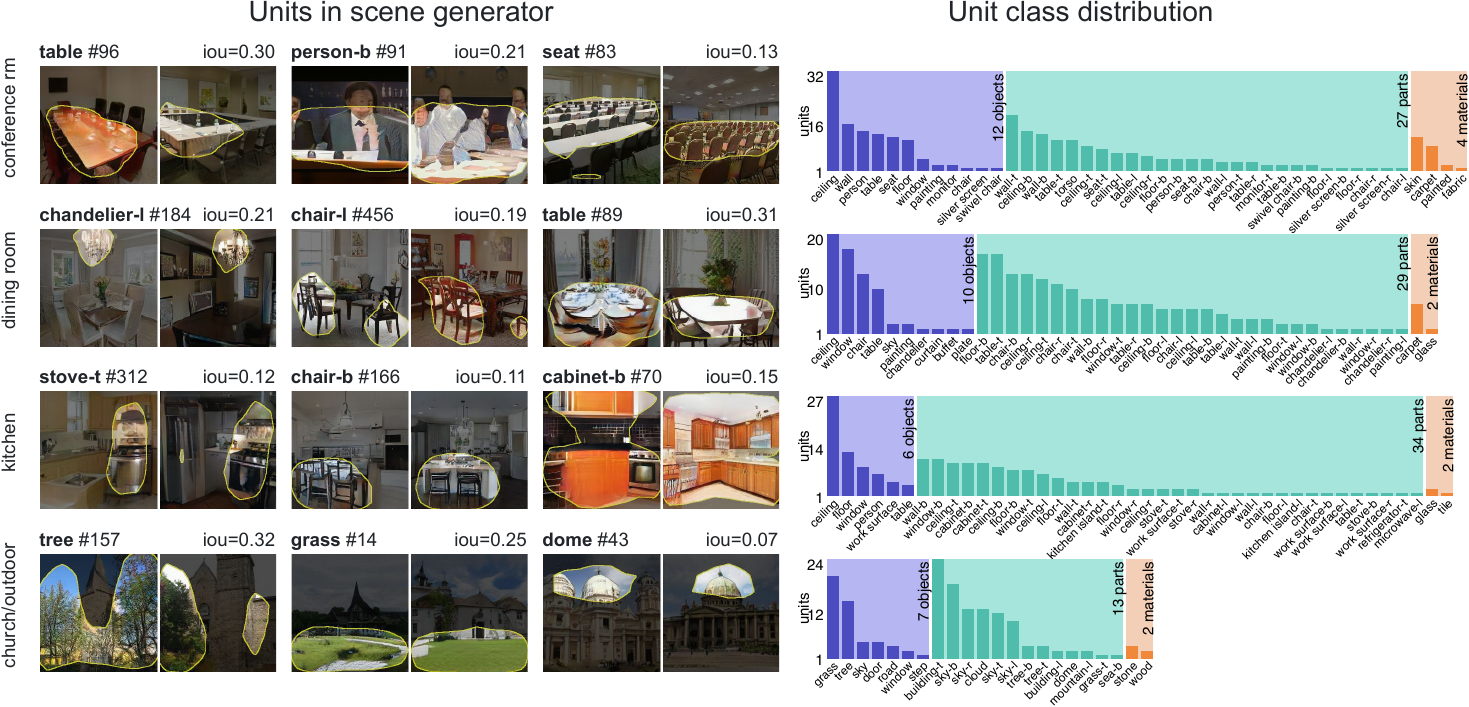}
\vspace{-20pt}
\caption{Comparing representations learned by progressive GANs trained on different scenes. Units match objects that commonly appear in the scene type, e.g., seats in conference rooms and stoves in kitchens.  A unit is counted as a class predictor if it matches a segmentation class with pixel accuracy $> 0.75$ and IoU $> 0.05$ when upsampled and thresholded. The distribution of units over classes is shown at right.}
\vspace{-5pt}
\lblfig{scene_units}
\end{figure}

\begin{figure}[t]
\includegraphics[width=\columnwidth]{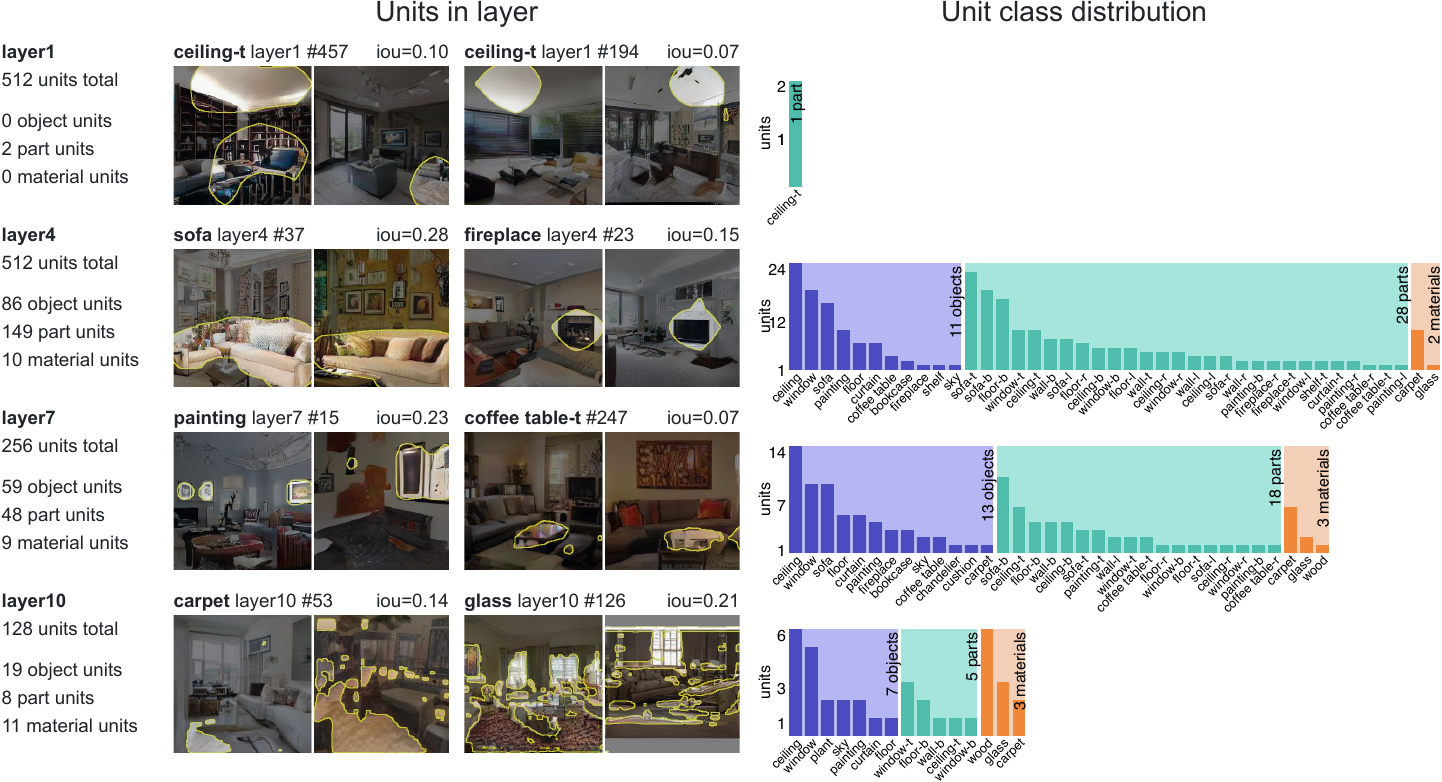}
\vspace{-20pt}
\caption{Comparing layers of a progressive GAN trained to generate $256\times256$ LSUN living room images. The output of the first convolutional layer has almost no units that match semantic objects, but many objects emerge at layers 4-7.  Later layers are dominated by low-level materials and shapes.}
\lblfig{compare_layers}
\end{figure}

\begin{figure}[t]
\includegraphics[width=\columnwidth]{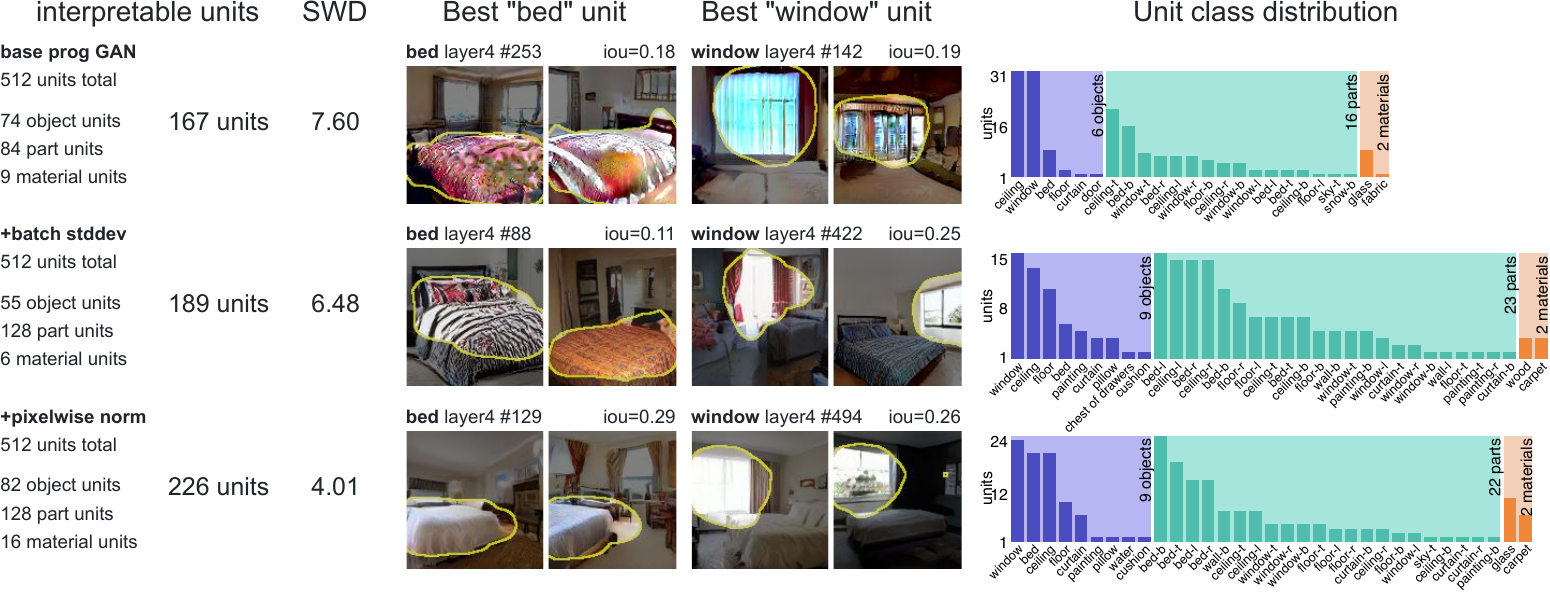}
\vspace{-15pt}
\caption{Comparing \layer{4} representations learned by different training variations. Lower SWD indicates a higher-quality model: as the quality of the model improves, the number of interpretable units also rises.  Progressive GANs apply several innovations including making the discriminator aware of minibatch statistics, and pixelwise normalization at each layer.  We can see batch awareness increases the number of object classes matched by units, and pixel norm (applied in addition to batch stddev) increases the number of units matching objects.}
\lblfig{compare_models}
\end{figure}
\begin{figure}[ht]
\includegraphics[width=\columnwidth]{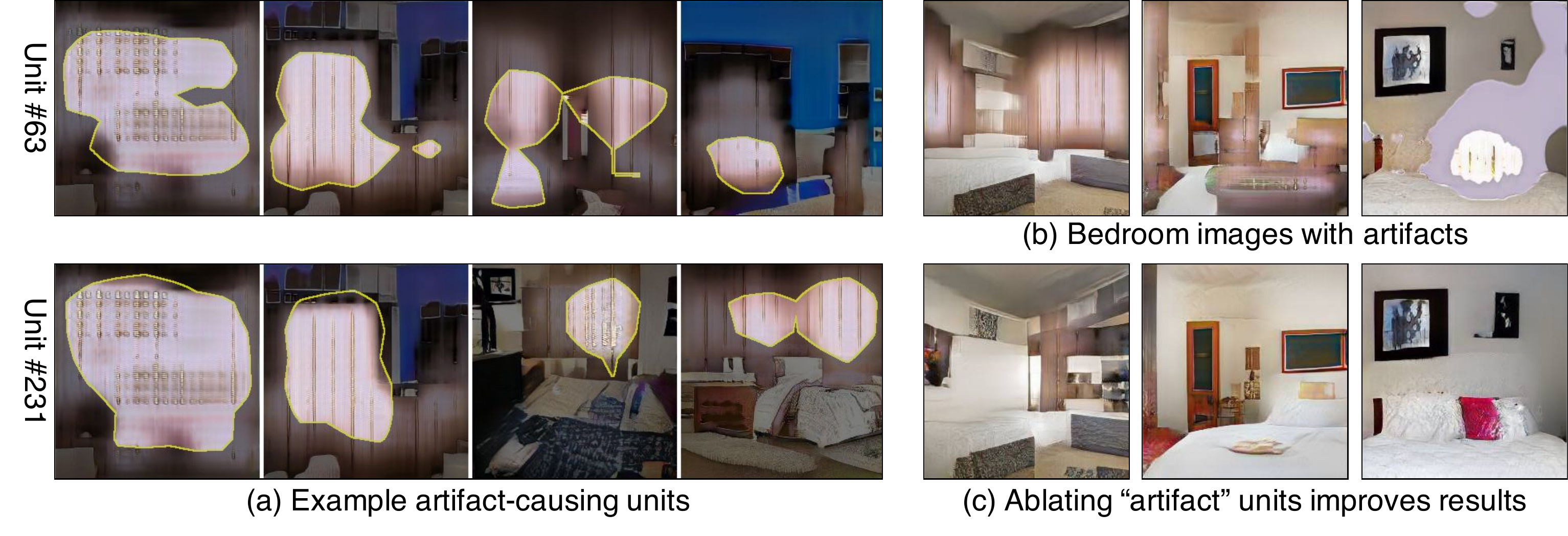}
\caption{(a) We show two example ``artifact''  units that are responsible for visual artifacts in GAN results. There are $20$ units in total. By ablating these units,  we can fix the artifacts in (b) and largely improve the visual quality as shown in (c). }
\lblfig{artifacts}
\end{figure}

\begin{table}[t]
	\small
	\centering
	\caption{We compare generated images before and after ablating $20$ ``artifacts'' units. We also report a simple baseline that ablates $20$ randomly chosen units.}

	\begin{tabularx}{175pt}{cc}
		\toprule
	 \multicolumn{2}{c}{\fid (FID)}	\tabularnewline\midrule
original images		  & 52.87    	\tabularnewline
		``artifacts'' units ablated (ours)		  & {\bf 32.11}   		\tabularnewline
		random units ablated      & 52.27
		\tabularnewline\bottomrule
	\end{tabularx}\quad
	\begin{tabularx}{210pt}{cc}
		\toprule
		Human preference score  & original images  \tabularnewline\midrule
		``artifacts'' units ablated (ours) 	  & {\bf 79.0}\% \tabularnewline
		random units ablated   &  50.8\% 
		\tabularnewline\bottomrule
	\end{tabularx}
	\lbltbl{artifacts}
\end{table}

\begin{figure}[t]
\centering
\includegraphics[width=0.95\columnwidth]{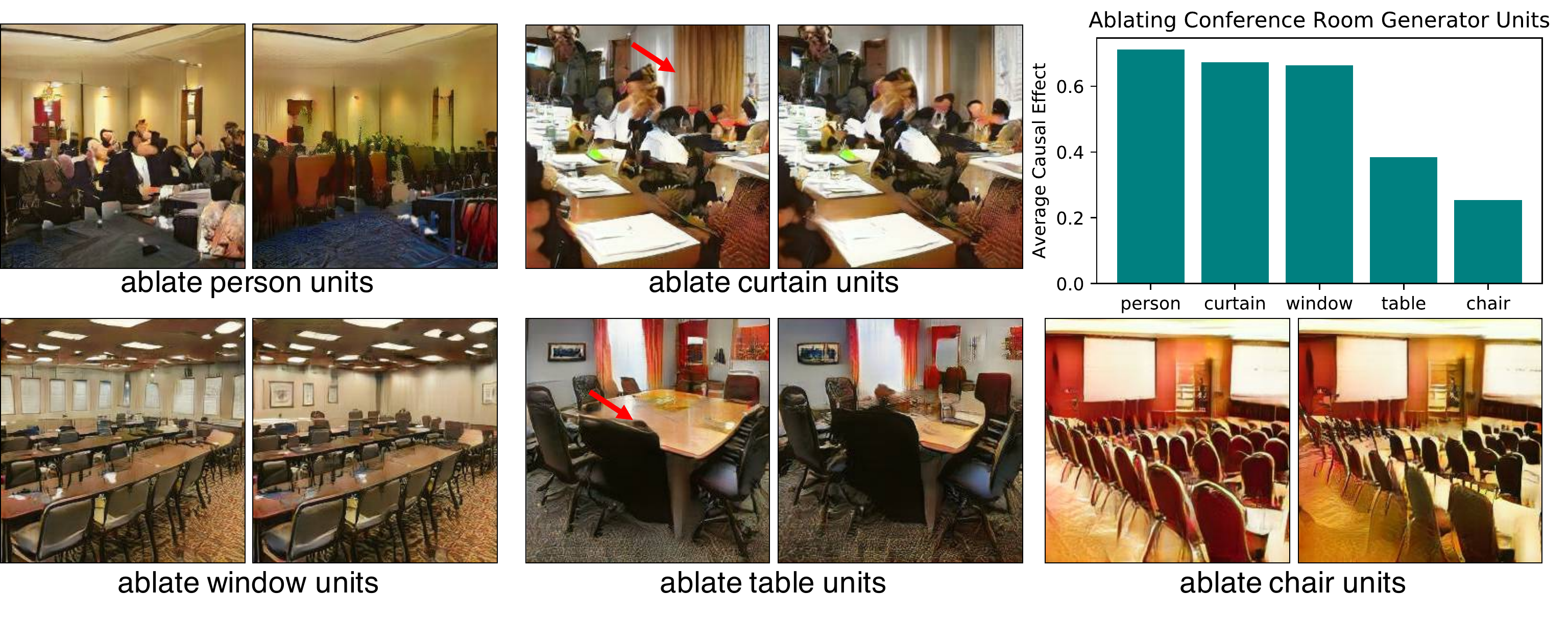}
\caption{Measuring the effect of ablating units in a GAN trained on conference room images. Five different sets of units have been ablated related to a specific object class.  In each case, $20$ (out of $512$) units are ablated from the same GAN model. The $20$ units are specific to the object class and independent of the image.  The average causal effect is reported as the portion of pixels that are removed in $1\,000$ randomly generated images.   We observe that some object classes are easier to remove cleanly than others: a small ablation can erase most pixels for people, curtains, and windows, whereas a similar ablation for tables and chairs only reduces object sizes without erasing them.}
\lblfig{ablation_confroom}
\end{figure}

\begin{figure}[t]
\centering
\includegraphics[width=\columnwidth]{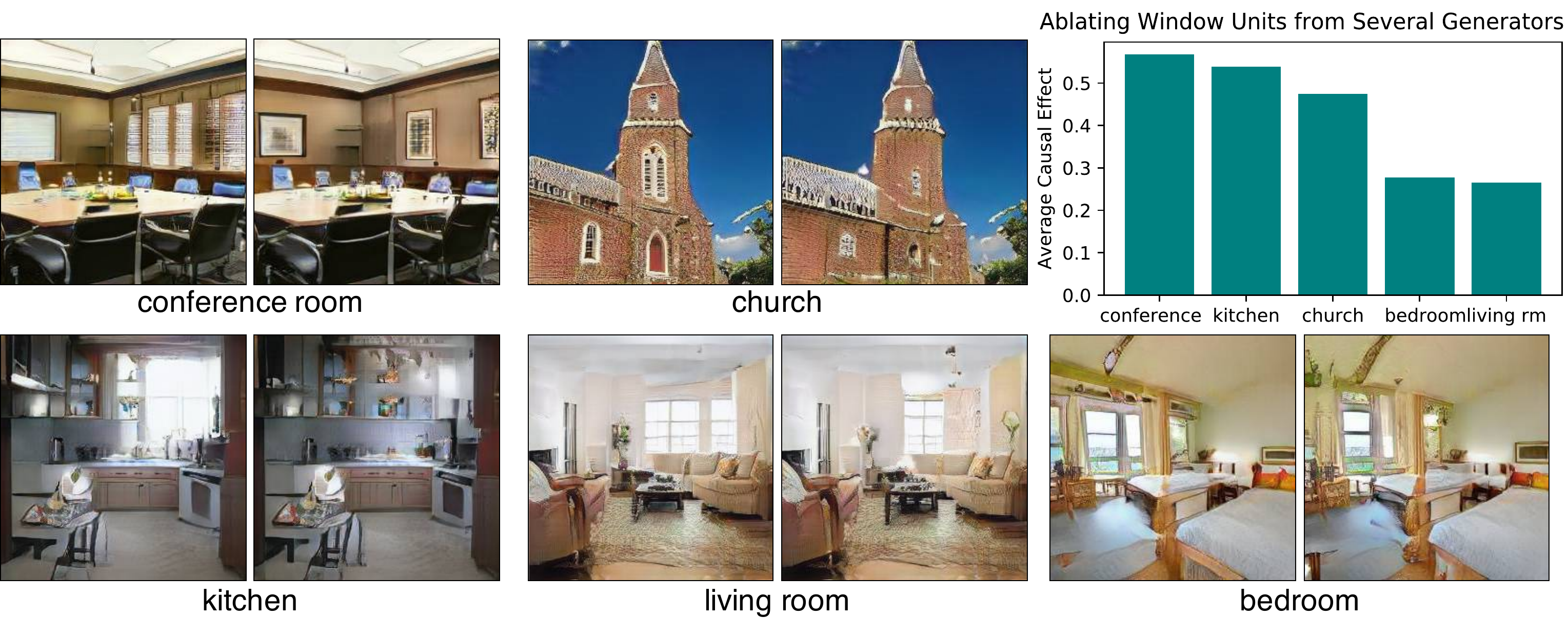}
\caption{Comparing the effect of ablating 20 window-causal units in GANs trained on five scene categories. In each case, the 20 ablated units are specific to the class and the generator and independent of the image.  In some scenes, windows are reduced in size or number rather than eliminated completely, or replaced by visually similar objects such as paintings.}
\lblfig{ablation_window}
\end{figure}

\begin{figure}
\centering
\includegraphics[width=0.95\columnwidth]{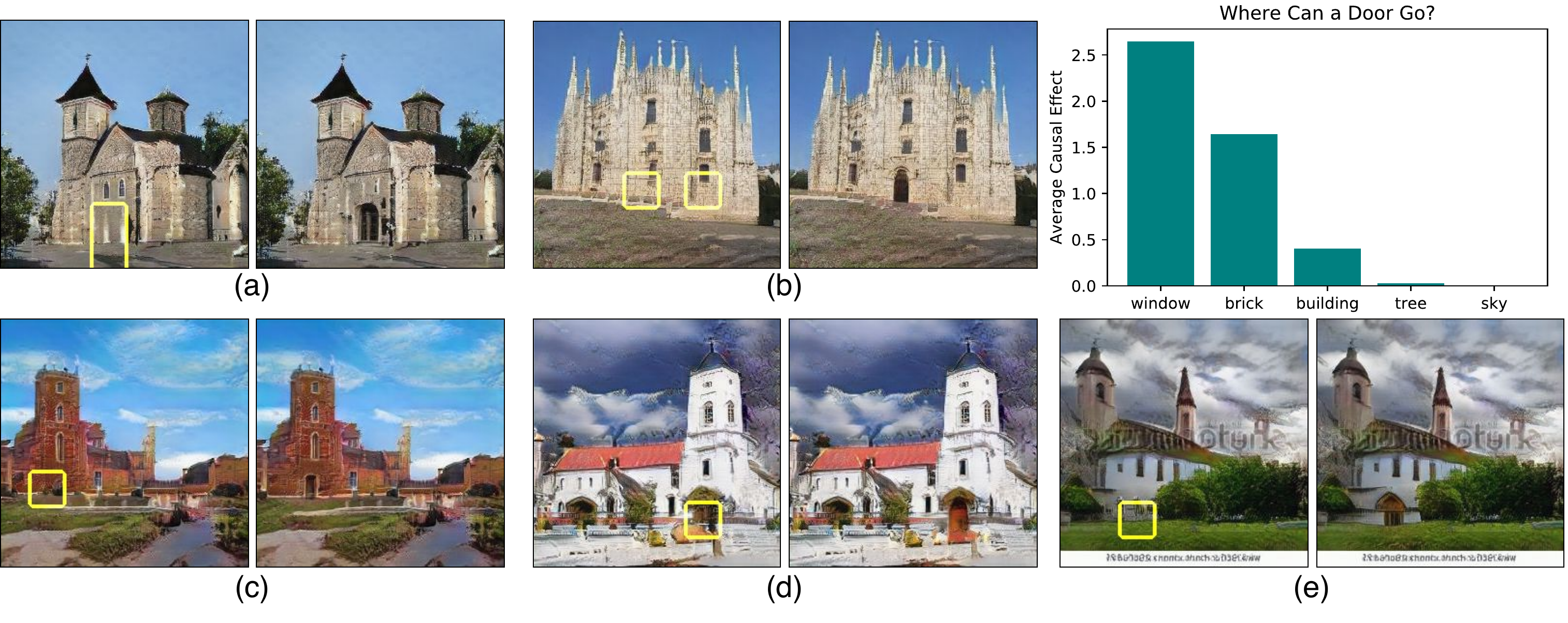}
\caption{
Inserting door units by setting $20$ causal units to a fixed high value at one pixel in the representation.  Whether the door units can cause the generation of doors is dependent on local context: every location that creates doors is shown, including two separate locations in (b) (we intervene at left). The same units are inserted in every case, but the door that appears has a size, alignment, and color appropriate to the location.  The top chart summarizes the causal effect of inserting door units at one pixel with different context. }
\lblfig{insertion}
\end{figure}

\noindent\textbf{This abstract is a record of an invited talk discussing work originally presented at ICLR 2019: GAN Dissection~\citep{bau2019gandissect}, arXiv at this location: \href{https://arxiv.org/abs/1811.10597}{https://arxiv.org/abs/1811.10597}.}

The ability of generative adversarial networks to render nearly photorealistic images leads us to ask: What does a GAN know?
For example, when a GAN generates a door on a building but not in a tree (\reffig{teaser}a), we wish to understand whether such structure emerges as pure pixel patterns without explicit representation, or if the GAN contains internal variables that correspond to human-perceived objects such as doors, buildings, and trees. And when a GAN generates an unrealistic image (\reffig{teaser}f), we want to know if the mistake is caused by specific variables in the network.

We present a method for visualizing and understanding GANs at different levels of abstraction, from each neuron, to each object, to the relationship between different objects. Beginning with a Progressive GAN~\citep{karras2018progressive} trained to generate scenes (\reffig{teaser}b), we first identify a group of interpretable units that are related to semantic classes (\reffig{teaser}a, \reffig{scene_units}). These units' featuremaps closely match the semantic segmentation of a particular object class (e.g., doors). Then, we directly intervene within the network to identify sets of units that cause a type of object to disappear (\reffig{teaser}c) or appear (\reffig{teaser}d). Finally, we study contextual relationships by observing where we can insert the object concepts in new images and how this intervention interacts with other objects in the image (\reffig{teaser}d, \reffig{insertion}). This framework enables several applications: comparing internal representations across different layers, GAN variants, and datasets (\reffig{scene_units}); debugging and improving GANs by locating and ablating artifact-causing units (\reffig{teaser}e,f,g); understanding contextual relationships between objects in natural scenes (\reffig{insertion} ,\reffig{layereffect}); and manipulating images with interactive object-level control (\href{http://tiny.cc/gandissect}{video}).

\section{Method}

We analyze the internal GAN representations by decomposing the featuremap $\repr$ at a layer into positions $\pixel \subset \pixelall$ and unit channels $u\in\Uall$.  To identify a unit $u$ with semantic behavior, we upsample and threshold the unit (\reffig{teaser}b), and measure how well it matches an object class $c$ in the image $\rvx$ as identified by a supervised semantic segmentation network $\seg(\rvx)$ \citep{xiao2018unified}
\begin{align}
\nonumber
\IoU_{u,c} &\equiv
\frac{\E_{\rvz} \left|(\uprepru > t_{u,c}) \land  \seg(\rvx) \right|}
{\E_{\rvz} \left|(\uprepru > t_{u,c}) \vee \seg(\rvx) \right|},
\end{align}
\begin{align}
\mbox{where } t_{u,c} &= \argmax_{t} \frac{\text{I}(\uprepru > t ; \seg(\rvx))}{\text{H}(\uprepru > t , \seg(\rvx))}
\label{eq:max-iqr}
\end{align}
This approach is inspired by the observation that many units in classification networks locate emergent object classes when upsampled and thresholded~\citep{bau2017network}.  Here, the threshold $t_{u,c}$ is chosen to maximize the information quality ratio, that is, the portion of the joint entropy $\textnormal{H}$ which is mutual information $\textnormal{I}$~\citep{wijaya2017information}.

To identify a sets of units $\U\subset\Uall$ that cause semantic effects, we intervene in the network $G(\rvz) = f(h(\rvz)) = f(\repr)$ by decomposing the featuremap $\repr$ into two parts $(\rUP, \rUPb)$, and forcing the components $\rUP$ on and off:

\vspace{\abovedisplayskip}\noindent Original image:
\begin{align}
\rvx = \G(\rvz)  \equiv  \f(\repr) & \equiv  \f(\rUP, \rUPb)
\end{align}
Image with $\U$ ablated at pixels $\pixel$:
\begin{align}
\xablate & =  \f(\vzero, \rUPb)
\end{align}
Image with $\U$ inserted at pixels $\pixel$:
\begin{align}
\xinsert & = \f(\thresU, \rUPb)
\end{align}
We measure the average causal effect (ACE) \citep{holland1988causal} of units $\U$ on class $c$ as:
\begin{equation}
\ACEUc  \equiv \Ezp[\seg(\xinsert)] - \Ezp[\seg(\xablate)],
\label{eq:ace}
\end{equation}

\section{Results}

\myparagraph{Interpretable units for different scene categories}
The set of all object classes matched by the units of a GAN provides a map of what a GAN has learned about the data. \reffig{scene_units} examines units from generators train on four LSUN~\citep{yu2015lsun} scene categories. The units that emerge are object classes appropriate to the scene type: for example, when we examine a GAN trained on kitchen scenes, we find units that match stoves, cabinets, and the legs of tall kitchen stools.  Another striking phenomenon is that many units represent parts of objects: for example, the conference room GAN contains separate units for the body and head of a person.

\myparagraph{Interpretable units for different network layers.}
In classifier networks, the type of information explicitly represented changes from layer to layer \citep{zeiler2014visualizing}. We find a similar phenomenon in a GAN.  \reffig{compare_layers} compares early, middle, and late layers of a progressive GAN with $14$ internal convolutional layers.  The output of the first convolutional layer, one step away from the input $z$, remains entangled.  Mid-level layers $4$ to $7$ have a large number of units that match semantic objects and object parts.  Units in layers $10$ and beyond match local pixel patterns such as materials and shapes.

\myparagraph{Interpretable units for different GAN models.}
Interpretable units can provide insight about how GAN architecture choices affect the structures learned inside a GAN. \reffig{compare_models} compares three models~\citep{karras2018progressive} that introduce two innovations on baseline \pgan.  By examining unit semantics, we confirm that providing minibatch stddev statistics to the discriminator increases not only the visible GAN output, but also the diversity of concepts represented by units of a GAN: the number of types of objects, parts, and materials matching units increases by more than $40\%$. The second architecture applies pixelwise normalization to achieve better training stability.
As applied to \pgan, pixelwise normalization increases the number of units that match semantic classes by $19\%$.

\myparagraph{Diagnosing and Improving GANs}
Our framework can also analyze the causes of failures in their results. \reffig{artifacts}a shows several annotated units that are responsible for typical artifacts consistently appearing across different images. Such units can be identified by visualizing ten top-activating images for each unit, and labeling units for which many visible artifacts appear in these images. Human annotation is efficient and it typically takes $10$ minutes to locate $20$ artifact-causing units out of $512$ units in \layer{4}. 

More importantly, we can fix these errors by ablating the $20$ artifact-causing units. \reffig{artifacts}b shows that artifacts are successfully removed and the artifact-free pixels stay the same, improving the generated results. To further quantify the improvement, we compute the \fid~\citep{heusel2017gans} between the generated images and real images using $50\,000$ real images and $10\,000$ generated images with high activations on these units. We also ask human participants on Amazon MTurk to identify the more realistic image given two images produced by different methods: we collected $20\,000$ annotations for $1\,000$ images per method. As summarized in \reftbl{artifacts}, our framework significantly improves fidelity based on these two metrics.

\myparagraph{Locating causal units with ablation}
Errors are not the only type of output that can be affected by directly intervening in a GAN.  A variety of specific object types can also be removed from GAN output by ablating a set of units in a GAN.  In \reffig{ablation_confroom} we intervene in sets of 20 units that have causal effects on common object classes in conference rooms scenes.  We find that, by turning off small sets of units, most of the output of people, curtains, and windows can be removed from the generated scenes.  However, not every object has a simple causal encoding: tables and chairs cannot be removed.  Ablating those units will reduce the size and density of these objects, but will rarely eliminate them.

The ease of object removal depends on the scene type.  \reffig{ablation_window} shows that, while windows can be removed well from conference rooms, they are more difficult to remove from other scenes.  In particular, windows are as difficult to remove from a bedroom as tables and chairs from a conference room.  We hypothesize that the difficulty of removal reflects the level of choice that a GAN has learned for a concept: a conference room is defined by the presence of chairs, so they cannot be removed.  And modern building codes mandate that bedrooms must have windows; the GAN seems to have noticed.

\myparagraph{Characterizing contextual relationships using insertion}
We can also learn about the operation of a GAN by forcing units on and inserting these features into specific locations in scenes.   \reffig{insertion} shows the effect of inserting $20$ \layer{4} causal door units in church scenes.  In this experiment, we insert units by setting their activation to the mean activation level at locations at which doors are present. Although this intervention is the same in each case, the effects vary widely depending on the context.  For example, the doors added to the five buildings in \reffig{insertion} appear with a diversity of visual attributes, each with an orientation, size, material, and style that matches the building.

We also observe that doors cannot be added in most locations. The locations where a door can be added are highlighted by a yellow box.  The bar chart in \reffig{insertion} shows average causal effects of insertions of door units, conditioned on the object class at the location of the intervention.  Doors can be created in buildings, but not in trees or in the sky.  A particularly good location for inserting a door is one where there is already a window.

\myparagraph{Tracing the causal effects of an intervention}
To investigate the mechanism for suppressing the visible effects of some interventions, we perform an insertion of 20 door-causal units on a sample of locations and measure the changes in later layer featuremaps caused by interventions at layer 4.  To quantify effects on downstream features, and the effect on each each feature channel is normalized by its mean L1 magnitude, and we examine the mean change in these normalized featuremaps at each layer.  In \reffig{layereffect}, these effects that propagate to \layer{14} are visualized as a heatmap: brighter colors indicate a stronger effect on the final feature layer when the door intervention is in the neighborhood of a building instead of trees or sky.  Furthermore, we graph the average effect on every layer at right in  \reffig{layereffect}, separating interventions that have a visible effect from those that do not.  A small identical intervention at \layer{4} is amplified to larger changes up to a peak at \layer{12}.

Interventions provide insight on how a GAN enforces relationships between objects.  We find that even if we try to add a door in \layer{4}, that choice can be vetoed by later layers if the object is not appropriate for the context.

\begin{figure}
\centering
\includegraphics[width=0.29\columnwidth]{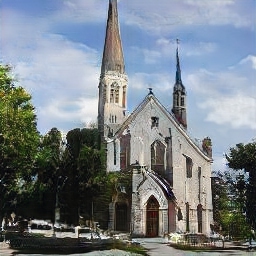}
\includegraphics[width=0.29\columnwidth]{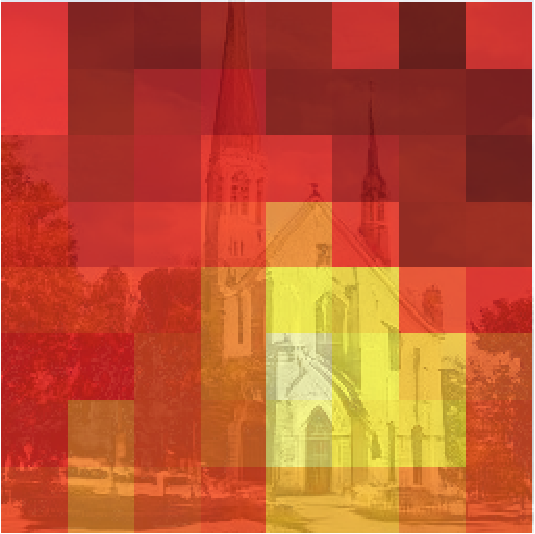}%
\includegraphics[width=0.4\columnwidth]{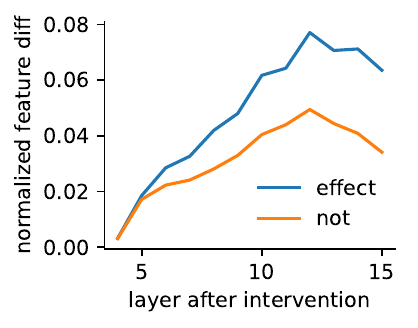}%
\caption{Tracing the effect of inserting door units on downstream layers. An identical "door" intervention at \layer{4} of each pixel in the featuremap has a different effect on final convolutional feature layer, depending on the location of the intervention.  In the heatmap, brighter colors indicate a stronger effect on the \layer{14} feature.  A request for a door has a larger effect in locations of a building, and a smaller effect near trees and sky.  At right, the magnitude of feature effects at every layer is shown, measured by mean normalized feature changes.  In the line plot, feature changes for interventions that result in human-visible changes are separated from interventions that do not result in noticeable changes in the output.
}
\lblfig{layereffect}
\end{figure}
\section{Discussion}
\lblsec{discussion}
By carefully examining representation units, we have found that many parts of GAN representations can be interpreted, not only as signals that correlate with object concepts but as variables that have a causal effect on the synthesis of semantic objects in the output.  These interpretable effects can be used to compare, debug, modify, and reason about a GAN model.

Prior visualization methods~\citep{zeiler2014visualizing,bau2017network,karpathy2016visualizing} have brought many new insights to CNN and RNNs research. Motivated by that, in this work we have taken a small step towards understanding the internal representations of a GAN, and we have uncovered many questions that we cannot yet answer with the current method. For example: why can't a door be inserted in the sky? How does the GAN suppress the signal in the later layers?  Further work will be needed to understand the relationships between layers of a GAN. Nevertheless, we hope that our work can help researchers and practitioners better analyze and develop their own GANs.

\small
\bibliography{reference.bib}
\bibliographystyle{aaai}

\end{document}